\documentclass[journal]{IEEEtran}

\usepackage{xcolor,soul,framed} 

\colorlet{shadecolor}{yellow}
\usepackage[pdftex]{graphicx}
\graphicspath{{../pdf/}{../jpeg/}}
\DeclareGraphicsExtensions{.pdf,.jpeg,.png}

\usepackage[cmex10]{amsmath}
\usepackage{array}
\usepackage{mdwmath}
\usepackage{mdwtab}
\usepackage{eqparbox}
\usepackage{url}
\usepackage{cite}
\usepackage{hyperref}

\usepackage{ragged2e} 
\usepackage{booktabs,makecell, multirow, tabularx}
\usepackage[ruled,linesnumbered]{algorithm2e}
\usepackage[numbers]{natbib}
\SetKwComment{Comment}{// }{}

\begin{document}
\bstctlcite{IEEEexample:BSTcontrol}
    \title{Towards Interactive and Learnable Cooperative Driving Automation: a Large Language Model-Driven Decision-Making Framework}
  \author{Shiyu Fang,
      Jiaqi Liu,
      Mingyu Ding,
      Yiming Cui,
      Chen Lv,~\IEEEmembership{Senior Member,~IEEE,} \\
      Peng Hang,~\IEEEmembership{Senior Member,~IEEE,} 
      and Jian Sun

\thanks{Corresponding author: Jian Sun}

}

\markboth{}{Roberg \MakeLowercase{\textit{et al.}}: High-Efficiency Diode and Transistor Rectifiers}

\maketitle

\begin{abstract}
Connected Autonomous Vehicles (CAVs) are being tested globally, but their performance in complex scenarios remains suboptimal. While cooperative driving improves CAV performance by leveraging vehicle collaboration, its lack of interaction and continuous learning limits current applications to single scenarios and specific Cooperative Driving Automation (CDA). To address these issues, this paper proposes CoDrivingLLM, an interactive and learnable LLM-driven cooperative driving framework for all-scenario and all-CDA applications. First, an environment module updates vehicle positions based on semantic decisions, mitigating errors from LLM-controlled positioning. Second, leveraging the four CDA levels defined in SAE J3216, a centralized-distributed coupled architecture reasoning module is developed to ensure safe and efficient cooperation through centralized negotiation and distributed decision. Finally, by introducing a memory module that employs Retrieval Augmented Generation (RAG), CAVs are endowed with the ability to learn from their past experiences to avoid repeating mistakes. Through ablation studies and comparisons with other cooperative driving methods, the results demonstrate that the proposed CoDrivingLLM significantly enhances safety, efficiency, and adaptability across various scenarios. Our code is available at: \url{https://github.com/FanGShiYuu/CoDrivingLLM}.

\end{abstract}



%
\IEEEpeerreviewmaketitle


\section{Introduction}
\IEEEPARstart{A}{s} autonomous driving technology continues to advance, we are entering an era where both Connected Autonomous Vehicles (CAVs) and Human-Driven Vehicles (HDVs) will coexist. While CAVs are considered to have great potential in improving traffic safety and efficiency, their current performance on open roads is far from satisfactory. According to California’s Department of Motor Vehicles \citep{DMV2019}, 51\% of disengagements were due to CAVs' decision-making failures. Additionally, the Beijing Autonomous Vehicle Road Test Report revealed that up to 91\% of disengagements occurred during interactions with other vehicles, indicating that current autonomous driving technology is not yet adequate for complex interaction scenarios \citep{beijing2022, liu2023real}.

To ameliorate this issue, a promising approach is to leverage the cooperative driving capabilities of CAVs \citep{stevens2017cooperative}. The SAE J3216 standard divides Cooperative Driving Automation (CDA) into four levels: Status-sharing, Intent-sharing, Agreement-seeking, and Prescriptive \citep{SAE2020}. Federal Highway Administration has conducted the CARMA Program to develop and test CDA. The SARTRE project, led by Volvo, has turned cooperative platooning into a reality. However, the current validation of the program primarily focuses on single CDA under single scenarios \citep{carma2022}. How to achieve all-scenario and all-CDA cooperative driving still remains to be tackled.


Various approaches have been proposed to address cooperative driving, typically categorized into optimization-based, rule-based, and machine learning-based methods. Optimization-based methods aim to optimize an objective function, often simplifying the problem to balance performance and computational efficiency. Common strategies include dividing the area into zones, each occupied by one vehicle \citep{yu2019managing, xu2019cooperative}, or decomposing the problem into two layers \citep{zhang2021trajectory, gao2022joint}. However, these methods often neglect driving regulations and social norms, making their decisions harder to interpret. In contrast, the rule-based approach boasts simplicity in form and thus computational efficiency \citep{chen2021connected, hu2021constraint}. \citet{chen2022conflict} conducted a comparison between the improved Depth-First Search Tree (iDFST) method in a large-scale intersection where over 30 vehicles were involved. Nevertheless, despite the rule-based approaches that can be coupled with traffic regulations and is suitable to large-scale systems, pre-set rules often result in poor robustness \citep{xu2018distributed}. On the other hand, machine learning methods, such as deep learning and reinforcement learning, have gained increasing popularity in recent years. These methods have been successfully applied to cooperative driving in environments such as intersections \citep{zhou2022cooperative, liu10315232}, merging zones \citep{hu2024cooperative, li2023simulation}, and highways \citep{chen2024communication}, demonstrating good performance. \citet{bae2020cooperation} combined RNN with MPC for cooperative driving in high-density scenarios, achieving strong real-time performance. \citet{zhou2022cooperative} utilized Multi-Agent Deep Deterministic Policy Gradient (MADDPG) to model the cooperation among agents and results showed that MADDPG outperforms IDM in fuel and efficiency by 7.4\% and 5.3\%. However, the performance of these learning-based models often degrades significantly when transferred to untrained environments \citep{kirk2023survey, ghosh2021generalization, xu2023decision}. Overall, while many studies have explored cooperative driving, the varying scene characteristics have led most research to focus on a single scenario. To achieve all-scenario and all-CDA, it is essential to establish a method that possesses both interactive and learnable capabilities.

Additionally, recent advancements in transformer models and Large Language Models (LLMs) have opened new possibilities for cooperative decision-making \citep{vaswani2023attentionneed, brown2020language}. These models have demonstrated strong potential in understanding human behavior and interacting with humans. \citep{Radford2019LanguageMA, wei2022chain, ouyang2022training}. \citet{sha2023languagempc} utilized LLMs’ understanding of common sense and complex scenarios, such as a malfunctioning vehicle in a roundabout, reducing overall costs by 18.1\% and 16.4\% compared to single-vehicle decision-making methods. Furthermore, to improve the interaction ability of CAV, research has introduced human instructions to establish LLM decision models aligned with human logic and results showed that Human-in-the-loop guidance for LLMs has significantly improved the interpretability of decisions \citep{zhang2024instruct, cui2024receive}. Building on this, \citet{cui2024personalized} introduced the Talk2Drive framework, a real-world deployment of LLMs, which reduced takeover rates by 56.6\% in various scenarios. 
To enable the LLM to learn from mistakes, \citet{wen2023dilu} introduced a reflection module to refine decisions using LLM knowledge, improving CAV learning. \citet{hu2024agentscodriver} introduced a communication module into LLM, which improved the success rate from 45\% to 75\%, highlighting the critical role of communication in enhancing CAV efficiency and safety in complex environments. Despite these advancements, most research still focuses on single-vehicle decision-making. Therefore, designing an efficient, safe, and socially compatible continuous learning framework based on LLMs for multi-vehicle cooperative driving tasks remains a significant area of exploration. 


Against this backdrop, given the significance of cooperative driving in addressing the current challenges faced by CAV in complex scenarios and the potential of LLM in enhancing the interaction and learning ability of CAV, this paper proposes an interactive and learnable LLM-driven cooperative driving framework for all-scenario and all-CDA. Our contributions can be summarized as follows:

\begin{itemize}
    \item This paper proposes CoDrivingLLM, a LLM-based interactive and learnable cooperative driving framework, which features a centralized-distributed coupled architecture. Through the effective design and coupling of the environment module, reasoning module, and memory module, the framework significantly enhances the performance of cooperative driving under different scenarios.
    \item In the reasoning module, four sub-modules including state-sharing, intent-sharing, negotiation, and decision are designed to enable flexible switching between different levels of CDA. Considering the high complexity of multi-vehicle cooperation, a conflict coordinator that infers the suggested passing order for vehicles is designed as the negotiation module. Results show that the introduction of the conflict coordinator significantly improves the vehicle interaction ability.
    \item By introducing the memory module with Retrieval Augment Generation (RAG), CAVs are endowed with the ability to continuously learn from historical experiences. By searching for memories most similar to the current scenario and referring to the corresponding decisions, CAVs can avoid repeating past mistakes. Results demonstrated that with an increasing number of historical interactions, the success rate of CoDrivingLLM steadily improved across different scenarios.
\end{itemize}

The rest of the paper is organized as follows. Section \uppercase\expandafter{\romannumeral2} formulates the problem of our research. Section \uppercase\expandafter{\romannumeral3} introduces the architecture of the proposed framework. In Section \uppercase\expandafter{\romannumeral4}, we validate our framework through several experiments. Finally, conclusions are made in Section \uppercase\expandafter{\romannumeral5}.

\section{Problem Formulation}
With the increasing maturity of autonomous driving technology, manufacturers have shifted from focusing on early-stage technical competition to prioritizing commercial deployment. However, CAVs still exhibit various problems on open roads and even become the culprit of many congestion or accidents. In certain scenarios, the accident rate of CAVs even reached 5.25 times that of human drivers \citep{abdel2024matched}, which has failed to meet people's expectations for autonomous driving technology, and is gradually eroding people's trust in CAVs. The communication capabilities of CAVs enable connectivity and mutual assistance. Therefore, leveraging cooperative driving capabilities is a promising way to enhance CAV performance.

Additionally, the cooperative decision-making problem for CAVs can be modeled as a Partially Observable Markov Decision Process (POMDP) \cite{spaan2012partially}. We define the POMDP using the tuple \( \mathcal{M}_{\mathcal{G}} = (\mathcal{V}, \mathcal{S}, [\mathcal{O}_i], [\mathcal{A}_i], \mathcal{P}) \), where:
\begin{itemize}
    \item \( \mathcal{V} \) represents the finite set of all controlled agents (CAVs),
    \item \( \mathcal{S} \) denotes the state space, encompassing the states of all agents and the environment,
    \item \( \mathcal{O}_i \) represents the observation space for each agent \( i \in \mathcal{V} \),
    \item \( \mathcal{A}_i \) denotes the action space for agent \( i \),
    \item \( \mathcal{P} \) represents the transition function, capturing the probability of moving from one state to another.
\end{itemize}

At any given time step, each agent \( i \) receives an individual observation \( o_i: \mathcal{S} \to \mathcal{O}_i \) and selects an action \( a_i \in \mathcal{A}_i \) based on a policy \( \pi_i : \mathcal{O}_i \times \mathcal{A}_i \to [0,1] \). The agent then transitions to a new state \( s_i^\prime \) with a probability given by the state transition function \( \mathcal{P}(s^\prime | s, a): \mathcal{S} \times \mathcal{A}_1 \times \cdots \times \mathcal{A}_N \to \mathcal{S} \), where \( N \) is the total number of agents.

\subsubsection{Observation Space}
 Due to the limitations of sensor hardware, a CAV can only detect the status information of surrounding vehicles within a limited distance $\mathcal{L}$. We denote the set of all observable vehicles within the perception range of agent \( i \) as \( \mathcal{N}_i \). 
 The observation space for agent \( i \), denoted as \( \mathcal{O}_i \), is a matrix with dimensions \( | \mathcal{N}_i | \times | \mathcal{F} | \), where \( |\mathcal{N}_i| \) represents the number of observable vehicles for agent \( i \), and \( | \mathcal{F} | \) represents the number of features used to describe a vehicle's state. The feature vector for vehicle \( i \) is expressed as:
\begin{equation}
\mathcal{F}_i = [x_i, y_i, v^x_i, v^y_i, lane_i,int_i]
\end{equation}
where \( x_i \) and \( y_i \) are the longitudinal and lateral positions, \( v^x_i \) and \( v^y_i \) are the longitudinal and lateral speeds, \( lane_i\) represents the lane information of vehicle \( i \), and \( int_i\) is the driving intention of vehicle \(i\). \(lane_i\) and \(int_i\) will be described in detail in the next section.

The overall observation space of the system is the combined observation of all CAVs, i.e., 
\[
\mathcal{O} = \mathcal{O}_1 \times \mathcal{O}_2 \times \cdots \times \mathcal{O}_{|\mathcal{V}|}.
\]

\subsubsection{Action Space}
Given that the strength of LLMs lies in their reasoning capabilities based on world knowledge rather than numerical computation, we design the decision actions of CAVs as discrete semantic decisions rather than direct vehicle control actions. The action space \( \mathcal{A}_i \) for agent \( i \) is defined as a set of high-level control decisions, including \( \{\textit{SLOWER}, \textit{IDLE}, \textit{FASTER}, \textit{LANE LEFT}, \textit{LANE RIGHT}\} \). Once a high-level decision is selected, lower-level controllers generate the corresponding steering and throttle control signals to manage the CAVs' movements. The overall action space is the combination of actions from all CAVs, i.e., 
\[
\mathcal{A} = \mathcal{A}_1 \times \mathcal{A}_2 \times \cdots \times \mathcal{A}_{|\mathcal{V}|}.
\]

\subsubsection{Transition Dynamics}
The transition dynamics of the system, encapsulated in the transition function \( \mathcal{P} \), are modeled by the kinematics of the vehicles. The motion of each CAV is influenced by both the decision actions of the agent and the behaviors of other surrounding vehicles, particularly HDVs in mixed traffic environments. The state update for each CAV is based on a combination of high-level semantic decisions and low-level control signals generated by the vehicle’s control system.

\begin{figure}[htbp]
  \begin{center}
  \centerline{\includegraphics[width=3.5in]{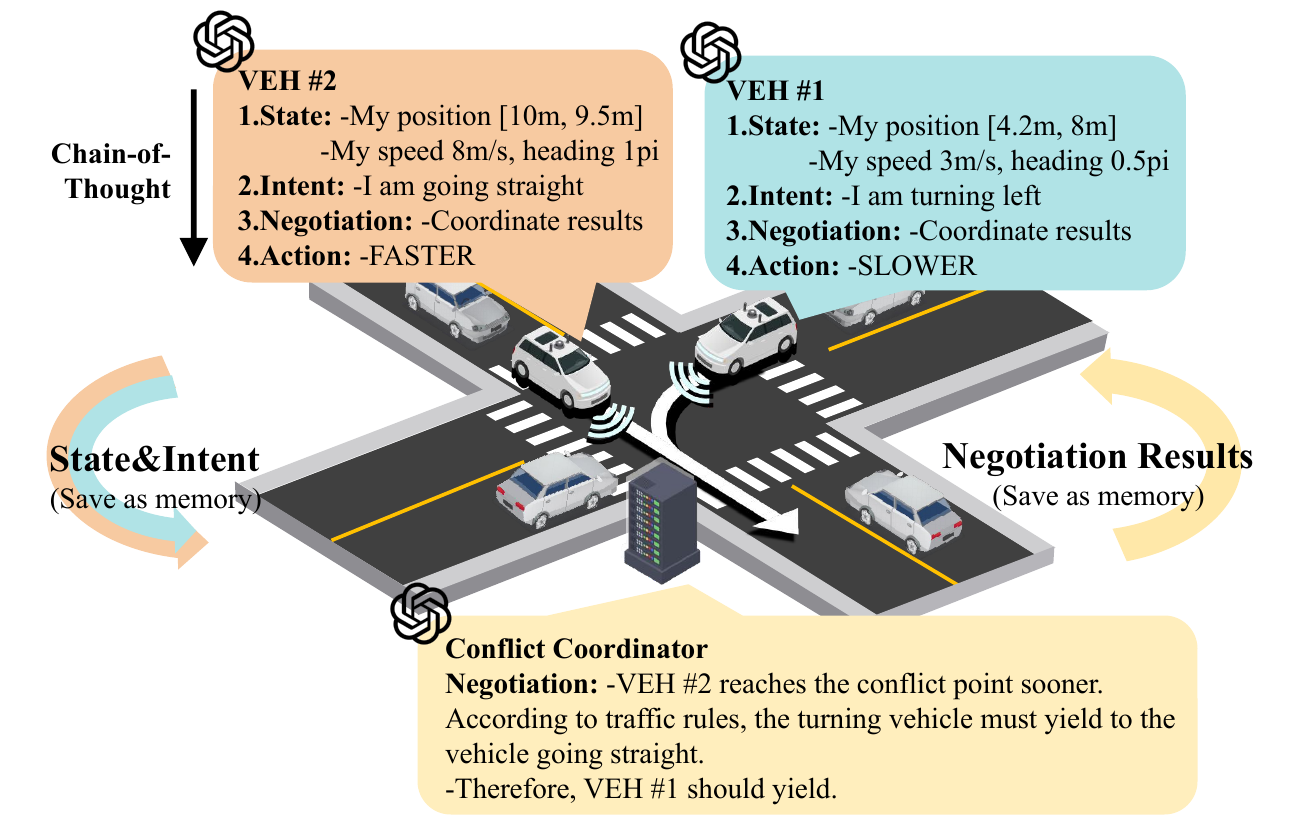}}
  \caption{Cooperative Driving with LLM.}\label{problem}
  \end{center}
  \vspace{-0.8cm}
\end{figure}

\section{Interactive and learnable LLM-driven cooperative driving framework}

Leveraging the cooperative driving ability of CAV can significantly improve the safety and efficiency of traffic systems. However, the current research often focuses on single-scenario, single-function cooperation, leading to insufficient interaction capabilities and a lack of continuous learning capabilities. Utilizing the extensive world knowledge and strong reasoning capabilities of LLMs holds promise for addressing these challenges and achieving all-scenario and all-CDA, thereby fully unleashing the potential of CAVs to transform traffic systems. As a solution, we propose CoDrivingLLM: an interactive and learnable LLM-driven cooperative driving framework for all-scenarios and all-CDA.

CoDrivingLLM mainly contains three modules: the environment module, the reasoning module, and the memory module. In this section, we first introduce the overall framework of CoDrivingLLM and then provide detailed explanations of each module.

\subsection{Overall Architecture}

\begin{figure*}[htbp]
  \begin{center}
  \centerline{\includegraphics[width=7in]{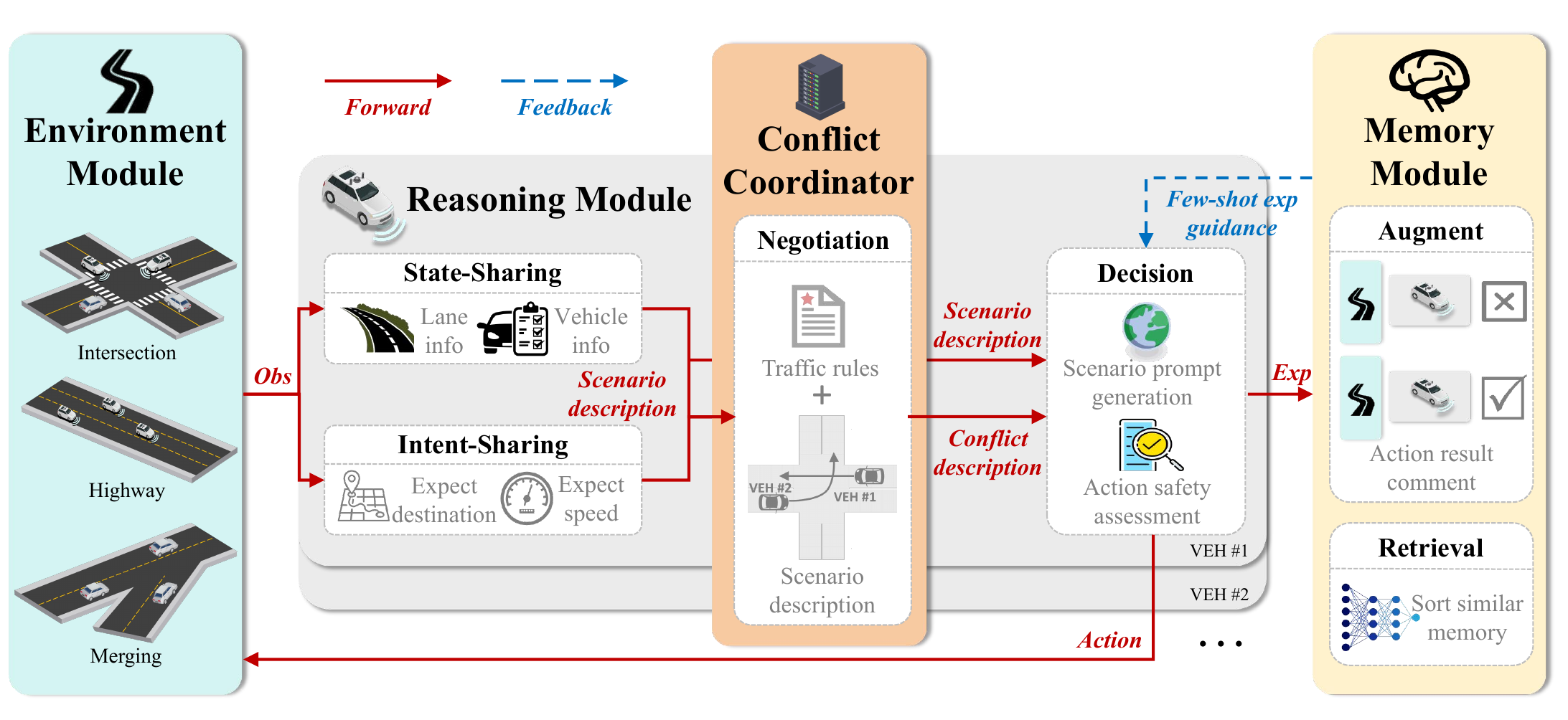}}
  \caption{The overview of CoDrivingLLM, an Interactive and Learnable LLM-Driven Cooperative Driving Framework.
  }\label{framework}
  \end{center}
  \vspace{-0.8cm}
\end{figure*}

Fig.~\ref{framework} illustrates the main modules and their logical relationships within CoDrivingLLM, which consists of three primary modules: the environment module, the reasoning module, and the memory module. First, the environment module updates the current scene information, including the states of all vehicles such as position, speed, etc., based on the actions of both CAVs and HDVs from the previous time step.
Next, we design a centralized-distributed coupled LLM reasoning module. Based on the four levels of CDA defined by the SAE J3216 standard, we integrate four sub-functions into this reasoning module: state sharing, intent sharing, negotiation, and decision. By incorporating the Chain-of-Thought (COT) method, we sequentially connect each sub-function in the reasoning process to enhance the safety and reliability of decision-making. During this process, each CAV uses LLM for distributed high-level logical reasoning and completes cooperative driving at various levels. Then, a conflict coordinator within the framework handles centralized conflict resolution, further improving safety. Finally, the scenario description, conflict description, and final decisions from the reasoning process are stored in a memory database with a vectorized form. In the subsequent reasoning, the CAV can reference the most similar past memory as experience, enabling the designed CAV to continuously learn and improve its capabilities while driving.


\subsection{Environment Module}
The environment module consists of two sub-modules: (1) the \textbf{environmental dynamics simulation sub-module}, which simulates real-world environmental dynamics to provide realistic background traffic flow and training feedback for the cooperative driving framework, and (2) the \textbf{model-based control execution sub-module}, which provides model-based vehicle control units for the LLM, enhancing the accuracy and success rate of action execution.

\subsubsection{Environmental Dynamics Simulation Sub-module}



Real-world feedback is crucial for training a stable and reliable cooperative decision-making model. To ensure the realism and reliability of the simulation, we consider a mixed human-autonomous driving environment within the environmental dynamics module, introducing uncontrolled human-driven vehicles to create a more realistic background traffic flow for the cooperative driving framework. These uncontrolled vehicles do not participate in cooperative driving tasks, meaning they do not share their intent and operate solely based on their own decision logic.

Given that the combination of IDM (Intelligent Driver Model) and MOBIL (Minimizing Overall Braking Induced by Lane changes) is widely used to characterize human driving behavior and has shown good results in various scenarios such as intersections, roundabouts, and merge areas, we utilize IDM and MOBIL to represent the longitudinal and lateral behaviors of HDVs, respectively. Since IDM and MOBIL are not the methodological contributions of this study, the detailed descriptions of these models can be found in the appendix here \footnote{\url{https://github.com/FanGShiYuu/CoDrivingLLM/blob/master/highway_env/HDV.md}}

\subsubsection{Model-based Control Execution Sub-module}



While LLMs possess strong reasoning capabilities, they do not perform well in precise mathematical calculations and low-level vehicle motion control. To address this, the target speed and target lane are adjusted based on the semantic output of the reasoning module. Subsequently, a model-based control approach is used to determine the acceleration and front wheel angle, which is formulated as follows:

\begin{equation}
\begin{split}
    a & = K_{p}(v_{r} - v) = K_{a}a_{r}\Delta t \\
    \delta & = \text{arcsin} (\frac{K_{h}(\phi_{r}-\phi)l}{2v})
    \label{control}
\end{split}
\end{equation}
where $K_{a}$ and $K_{h}$ represent the proportional gains for longitudinal control and lateral control, respectively, with values set to 1.6 and 5, $v_{r}$ and $a_{r}$ are the reference velocity and reference acceleration, $\phi_{r}$ is reference heading to follow the lane, and $\phi$ is the heading. 

After obtaining the vehicle's acceleration and front wheel angle from the control module, a bicycle model is selected as the kinematic module to generate the states of all vehicles in the environment at the next time step. The specifics are as follows:

\begin{equation}
f(a, \delta)=
\left \{
\begin{aligned}
    &x=v \cos(\phi+\beta) \\
    &y=v \sin(\phi+\beta) \\
    &v=a \\
    &\phi =\frac{v}{l}\sin\beta  \\
    &\beta=\tan^{-1}(\frac{1}{2\tan\delta}), 
    \label{kinematic}
\end{aligned}
\right
.
\end{equation}
where [x, y] is the vehicle position, $\beta$ is the slip angle at the center of gravity.

According to the aforementioned control module and kinematic module, the environment module can update the scenario information for the next time step according to CAV's semantic decision. During this approach, the vehicle positions are calculated using precise mathematical formulas, thereby avoiding the potential uncertainties and errors that might arise from LLM directly controlling the vehicle.

\subsection{Reasoning Module}




In this subsection, we establish an integrated reasoning module that progresses from state sharing to intent sharing, negotiation, and finally, decision. This module operates in a Chain-of-Thought (CoT) manner, ensuring a smooth transition for CAVs from environmental perception to interaction and negotiation, and ultimately to decision-making.

The reasoning module first extracts information about surrounding vehicles from the environment to create a scene description. It then organizes the states of the vehicles into conflict pairs, forming conflict descriptions. To ensure consistency in vehicle decision-making during conflicts and to avoid collisions, we develop an LLM-based conflict coordinator. This coordinator integrates the current conflict descriptions with traffic rules to determine the order of priority for each conflict group. Finally, each CAV makes its decision based on the conflict coordinator's recommendations and its own scenario description.


\subsubsection{State-sharing}



The state-perception function is responsible for acquiring and processing information about the current environment of the CAV, including dynamic data such as lane information and vehicle information. The designed state-sharing function aligns with the first level of CDA in the SAE J3216 standard, which is state sharing. CAVs are allowed to exchange information with others, therefore paving the way for subsequent higher levels of CDA. Through comprehensive analysis of the above information, the state-sharing function can construct a complete and accurate driving environment recognition, which provides a reliable basis for subsequent reasoning.

Specifically, lane information is divided into three categories: the ego lane, adjacent lanes, and conflict lanes. The ego lane is the lane where the ego vehicle is currently driving in, adjacent lanes are the lanes on the left and right sides of the ego lane if they exist, and conflict lanes are the lanes that intersect with the ego lane. Similarly, vehicle information can be grouped based on their relationship with the ego vehicle into leading vehicles, rearing vehicles, and conflict vehicles. During the state-sharing process, lane information and vehicle information are combined to create an overview of the ego vehicle's surrounding environment. Since vehicles in different lanes can affect the ego vehicle differently, a three-level action safety assessment is designed to ensure safe driving, which will be detailed in the decision subsection.

\subsubsection{Intent-sharing}

The intent-sharing function, which conveys the driving intentions of a vehicle to other CAVs, is a key advantage of cooperative driving. From macro to micro, driving intention mainly includes the sharing of expected lane and expected speed. Through intent-sharing, other vehicles can better understand the ego vehicle's intentions, allowing them to make decisions while avoiding conflicts as much as possible.

State sharing and intent sharing are combined as the scene descriptions. This description is based on the lane and includes all objects that could affect the ego vehicle, such as adjacent vehicles and conflicting vehicles. Therefore, it can be migrated to multiple scenarios, like intersections, highways, and merging areas, significantly enhancing the versatility of the proposed cooperative driving framework.

\subsubsection{Negotiation}



CAVs possess broader perception capabilities and more powerful computing capabilities, which theoretically enable faster and more accurate recognition of the intentions of other vehicles. However, according to California’s Department of Motor Vehicles \citep{DMV2022}, over 31\% of accidents are caused by CAV misjudging the intentions of other traffic participants, highlighting that traditional decision-making methods still have significant shortcomings in processing high-dimensional information such as intentions. Therefore, a conflict coordinator is designed to resolve conflicts and achieve level-3 CDA, which is known as agreement-seeking cooperation.

The conflict coordinator identifies all potential conflicts in the current environment and assesses the severity of each conflict based on the current state of both vehicles involved. To quantify the degree of conflict severity, the time difference to the conflict point as the surrogate indicator, defined as:
\begin{equation}
\begin{aligned}
\Delta TTCP& =\begin{vmatrix}TTCP_i-TTCP_j\end{vmatrix} = \begin{vmatrix}\frac{d_{i}}{v_{i}}-\frac{d_{j}}{v_{j}}\end{vmatrix} \\
&=\begin{cases}
\Delta TTCP\leq2s, \quad \text{Serious Danger}\\
2s<\Delta TTCP\leq5s, \quad \text{General Danger}\\
5s<\Delta TTCP\leq8s, \quad \text{Slight Danger}\\
\Delta TTCP\geq8s, \quad \text{No Danger}
\end{cases}
\end{aligned}
\label{ttcp}
\end{equation}
where $TTCP$ is the time to conflict point based on the vehicle's current distance to conflict point $d$ and speed $v$. Among them, when $\Delta TTCP$ is less than 2s, it signifies a serious conflict, requiring at least one of them must take an emergency brake. If $\Delta TTCP$ falls between 2s and 5s, it is considered that there is a general conflict between the two vehicles, and at least one vehicle should slow down and yield. When the $\Delta TTCP$ is between 5s and 8s, the vehicles are considered to have a slight collision, and vehicles should not accelerate simultaneously. When $\Delta TTCP$ is greater than 8s, it is considered safe that the conflict will not cause a collision. After discretizing $\Delta TTCP$ into several intervals, the different action risk is described in prompt tailored to the severity of the conflict.



The conflict coordinator determines the passing sequence of vehicles in each conflict pair based on the severity of the conflict. Based on the generated scenario description, information such as the distance to the conflict point and the time to the conflict point is derived to characterize the conflict description. In this process, the conflict coordinator utilizes the conflict description while simultaneously considering traffic rules and social norms in driving to produce the negotiation result. For example, as shown in Fig.~\ref{problem}, according to traffic rules, turning vehicles should yield to going-straight vehicles. Hence, after negotiation, the conflict coordinator determines that $\text{CAV}_2$ should yield in this pair of conflicts. The negotiation result and reasons are sent to the decision function to make the final decision. However, it is important to note that the negotiation result is advisory, and the final decision also depends on other factors in the ego vehicle's surrounding environment.

\subsubsection{Decision}



The decision module synthesizes the information from the three aforementioned modules to generate the final decision. Considering that the strength of LLMs lies in their reasoning capabilities based on world knowledge rather than numerical computation, we design the decision actions as discrete semantic decisions. The action space $A$ is designed as a set of high-level meta-actions. These meta-actions are then mapped to specific reference acceleration $a_{r}$ and reference heading $\phi_{r}$ values through based on control model. Moreover, the vehicle's next state is determined based on Eq.~\ref{control} and Eq.~\ref{kinematic} to ensure safe and stable control.

The decision function mainly includes two tasks: scenario prompt generation and action safety assessment. Scenario prompt generation involves integrating the scene descriptions produced by state sharing and intent sharing, as well as the negotiation results from the conflict coordinator, to summarize the key information in the surrounding driving environment. This integration serves as the prompt for LLM decision-making. Furthermore, compared with traditional decision-making methods, the safety of LLM decisions is often difficult to guarantee. Therefore, a three-layer action safety assessment method is proposed to determine whether action would pose safety risks in the current driving environment. 

\begin{figure}[htbp]
  \begin{center}
  \centerline{\includegraphics[width=3.5in]{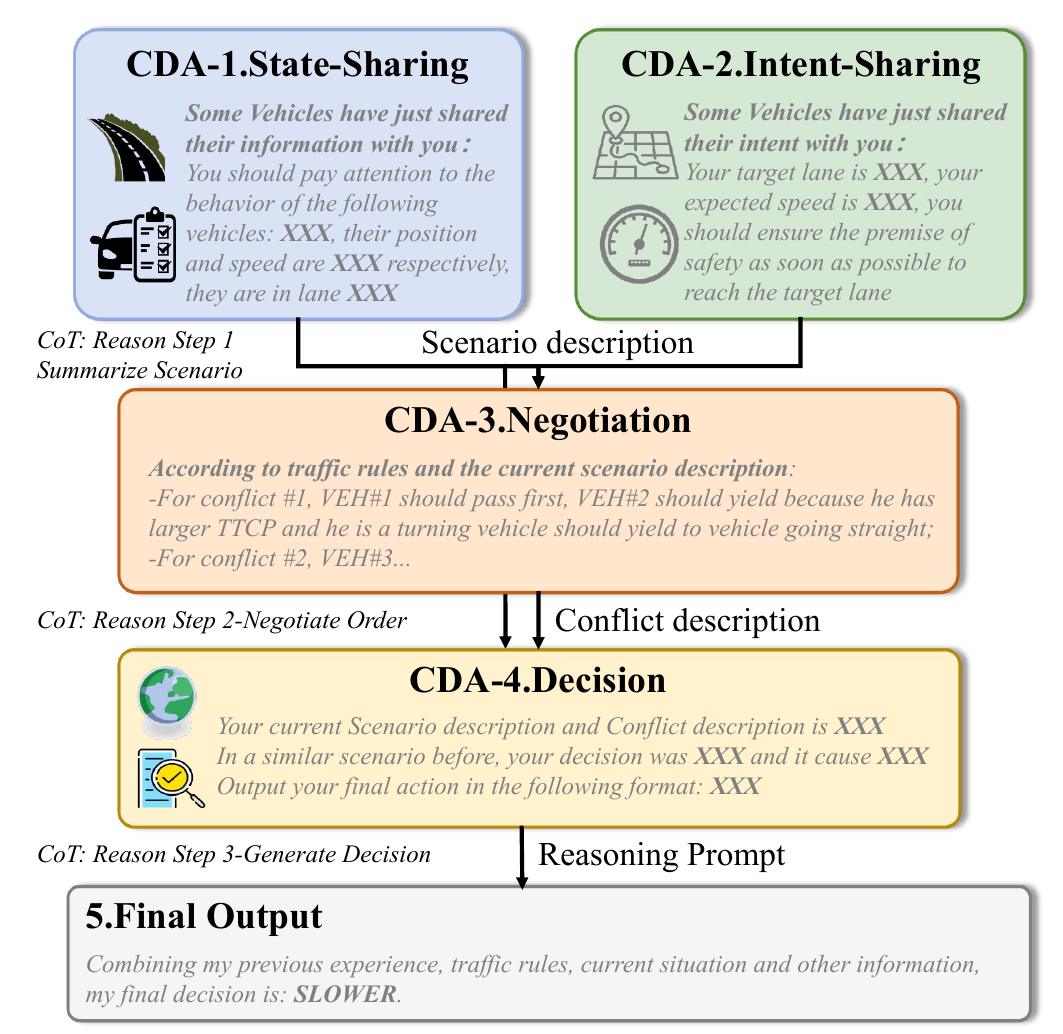}}
  \caption{Prompt for LLM Reasoning Module.}\label{prompt}
  \end{center}
  \vspace{-0.8cm}
\end{figure}

Based on lane classification, the first layer check determines whether the action will cause collisions with vehicles in the same lane, such as excessive acceleration leading to rear-end collisions or excessive deceleration causing abrupt braking by the rearing vehicle. Secondly, when the ego vehicle intends to take a left or right lane change, check whether the lane-changing will collide with other vehicles in the adjacent lane. Finally, check whether the action will lead to an intensification of the conflict with the vehicle in the conflict lane, especially avoiding serious and general dangers. During action safety assessment, any action found to have safety issues in the above checks is removed from the alternative action set. Therefore, by systematically assessing the feasibility of each action, the safety of LLM's final decision is significantly improved. Fig.~\ref{prompt} summarizes the composition of LLM's prompt in the inference process with CoT.

\subsection{Memory Module}

Enhancing continuous learning capabilities in autonomous systems has always been a significant challenge. Novice drivers accumulate experience through continuous driving practice, evaluate the effects of different behaviors, and learn from them to improve their driving skills. Drawing on this mechanism, a memory module is introduced to enable CAVs to learn from past experiences and utilize this knowledge to future interactions. This process is also referred to as RAG.

RAG endows LLMs with the ability to access specific knowledge databases within a domain or organization. This capability allows for economically efficient improvement of LLM outputs without requiring model retraining, ensuring relevance, accuracy, and practicality in addressing specific domain problems. Specifically, the designed memory module contains two primary functions: memory augment and memory retrieval.

\subsubsection{Memory augment}
The memory augment function evaluates the impact of CAV actions from the previous scenario to determine whether these actions have exacerbated conflicts. If a CAV's behavior leads to increased danger, the system generates negative feedback, such as: "Your action has intensified the conflict; similar actions should be avoided." This feedback mechanism establishes a connection between scenarios, actions, and results, storing these mappings in the memory databases for future reference. Before each invocation of the LLM for reasoning, the most relevant memories are retrieved from the memory databases to augment the prompts, thereby avoiding the repetition of past mistakes.

\subsubsection{Memory retrieval}
As the number of interactions increases, the memory databases will accumulate numerous past experiences. Inputting all memories as prompts would lead to redundancy, making it difficult for the CAV to extract key information during inference. To address this problem, a memory retrieval function is employed to extract the most relevant memories to the current scenario from the databases before using memory to guide reasoning.

Specifically, current scenario descriptions and conflict descriptions are converted into vector forms and cosine similarity is employed to rank memories in the library based on their relevance to the current scenario. The top-ranked memories are then selected as part of the prompts for CAV inference. These similar memories, referred to as few-shot experiences, are injected into the CAV's reasoning module, enabling the CAV to learn from past mistakes. 

The introduction of the memory module not only improves the CAV's decision-making ability in complex environments but also imparts human-like continuous learning capabilities. By continually learning from past experiences, CAVs can better adapt to dynamic environments, enhancing driving safety, reducing traffic accidents, and increasing the reliability and practicality of their real-world applications.

Algorithm.~\ref{algorithm} summarizes the main process of our framework. In the next section, we verify the effectiveness of our framework in different scenarios through simulation experiments.

\begin{algorithm}[h]
\caption{LLM-driven cooperative driving framework.}\label{algorithm}
\KwIn{Vehicle states $\boldsymbol{S}_{t}$, CAV list $\boldsymbol{C}$}
\KwOut{Vehicle next states $\boldsymbol{S}_{t+1}$}
\Comment{Reasoning Module}
Initialize the decision buffer of each CAV $\boldsymbol{D} \leftarrow []$\;
Initialize the scenario description buffer of each CAV $\boldsymbol{sce} \leftarrow []$\;
\ForEach{$i \in \boldsymbol{C}$}{
Generate the scenario description $sce_{i}$ of vehicle $i$ based on current states $\boldsymbol{S}_{t}$\;
Add scenario description to buffer $\boldsymbol{sce} \leftarrow \boldsymbol{sce}.\text{Add}(sce_{i})$ \;
}
Generate the conflict description $\boldsymbol{con}$ with conflict coordinator based on scenario description $\boldsymbol{sce}$ \;
\ForEach{$i \in \boldsymbol{C}$}{
Sort the conflict description related to the ego vehicle $con_{i} \leftarrow \boldsymbol{con}.\text{Sort}(i)$\;
\Comment{Memory Module}
Retrieve the most relevant memories $m$ to the current scenario $sce_{i}$ and conflict $con_{i}$ \;
Generate the prompt of decision based on memories $m$, scenario description $sce_{i}$, and conflict description $con_{i}$ \;
Reasoning the final semantic decision $d_{i}$ \;
Add decision to buffer $\boldsymbol{D} \leftarrow \boldsymbol{D}.\text{Add}(d_{i})$ \;
}
\Comment{Environment Module}
Update vehicle next state $\boldsymbol{S}_{t+1}$ with decision buffer $\boldsymbol{D}$ based on Eq.~\ref{control} and Eq.~\ref{kinematic}\;
Evaluate the impact of the decision and add to the memory database\;
\end{algorithm}

\section{Experiments and Analysis}
To verify whether the proposed CoDrivingLLM can effectively improve the interaction ability and learning ability of CAV, this section carries out verification from three aspects. Firstly, ablation experiments were conducted on the various sub-module in Reasoning Module in different scenarios. Second, the learning ability of CoDrivingLLM was evaluated by comparing its performance across different scenarios after experiencing varying numbers of interactions. Finally, we evaluated the safety and efficiency of CoDrivingLLM against other cooperative driving methods, including optimization-based, rule-based, and machine learning-based methods.

\subsection{Experiment Settings}
\begin{figure}[!htbp]
    \centering
    \includegraphics[width=3.2in]{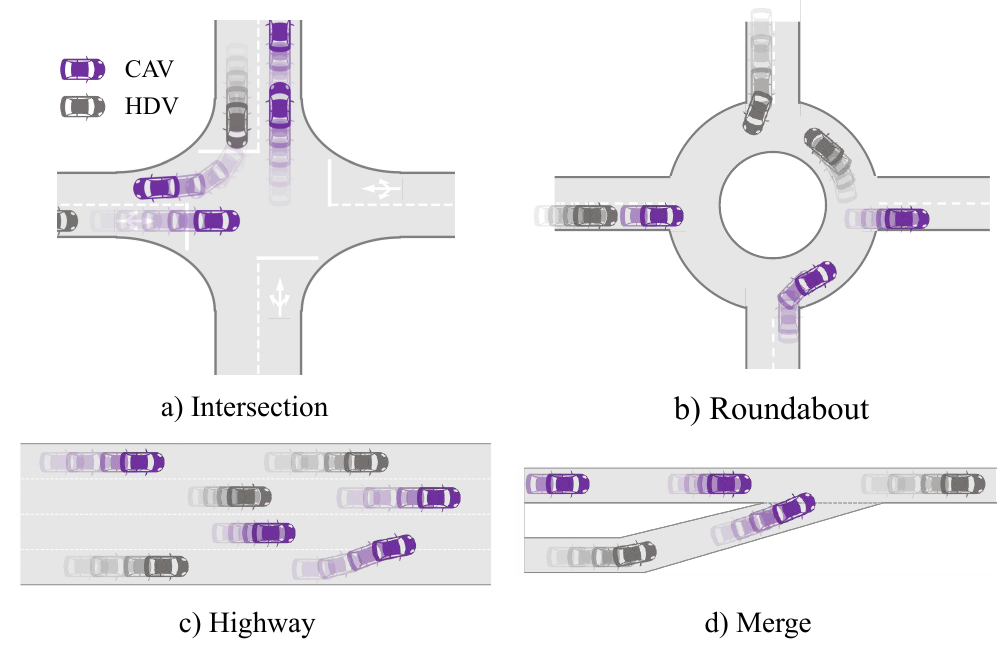}
    \caption{Four scenarios we utilized to train and test our method, (a) scenario 1, intersection environment; (b) scenario 2, roundabout environment; (3) scenario 3, highway environment; (4) scenario 4, merge environment.}
    \label{fig:scenario_setting}
\end{figure}

\textbf{Simulation Environment.} We develop our environment module based on the highway-env \cite{highway-env}, an open-sourced and flexible simulation platform for autonomous driving. Four scenarios are designed to conduct experiments, as shown in Fig.\ref{fig:scenario_setting}, including single-lane unsignalized intersection, roundabout scenario, four-way highway scenario, and merge scenario. The settings of context traffic flow and HDVs follow the instructions of the environment module of our framework.

\textbf{Implementation details.} The GPT-4o mini is utilized as our base LLM model to conduct high-level logical thinking and judgment. 
Furthermore, all scenarios are repeated 20 times using different random seeds to randomly generate the initial positions, speeds, and expected destinations of the vehicles. The success rate is used as the indicator to evaluate the performance of all methods. A case is considered successful if all CAVs finish their tasks safely and reach their destination.

\subsection{Ablation Study on Sub-modules of the Reasoning Module}

Cooperative driving has the potential to enhance the performance of CAVs in complex scenarios. Following the SAE J3216 standard, we developed a reasoning module that integrates state sharing, intention sharing, negotiation, and decision. To validate the effectiveness of these sub-modules, ablation studies were conducted in this subsection.

When the state-sharing sub-module is disabled, CAVs can only access the position and speed information of nearby vehicles within a 50-meter range. Moreover, CAVs are unable to detect potential conflicts between other vehicles and themselves without the intention-sharing sub-module. Finally, when the negotiation sub-module is absent, each CAV independently generates its expected passing order based on its understanding of the scenario information, rather than relying on a centralized conflict coordinator.
\begin{figure}[htbp]
  \begin{center}
  \centerline{\includegraphics[width=3in]{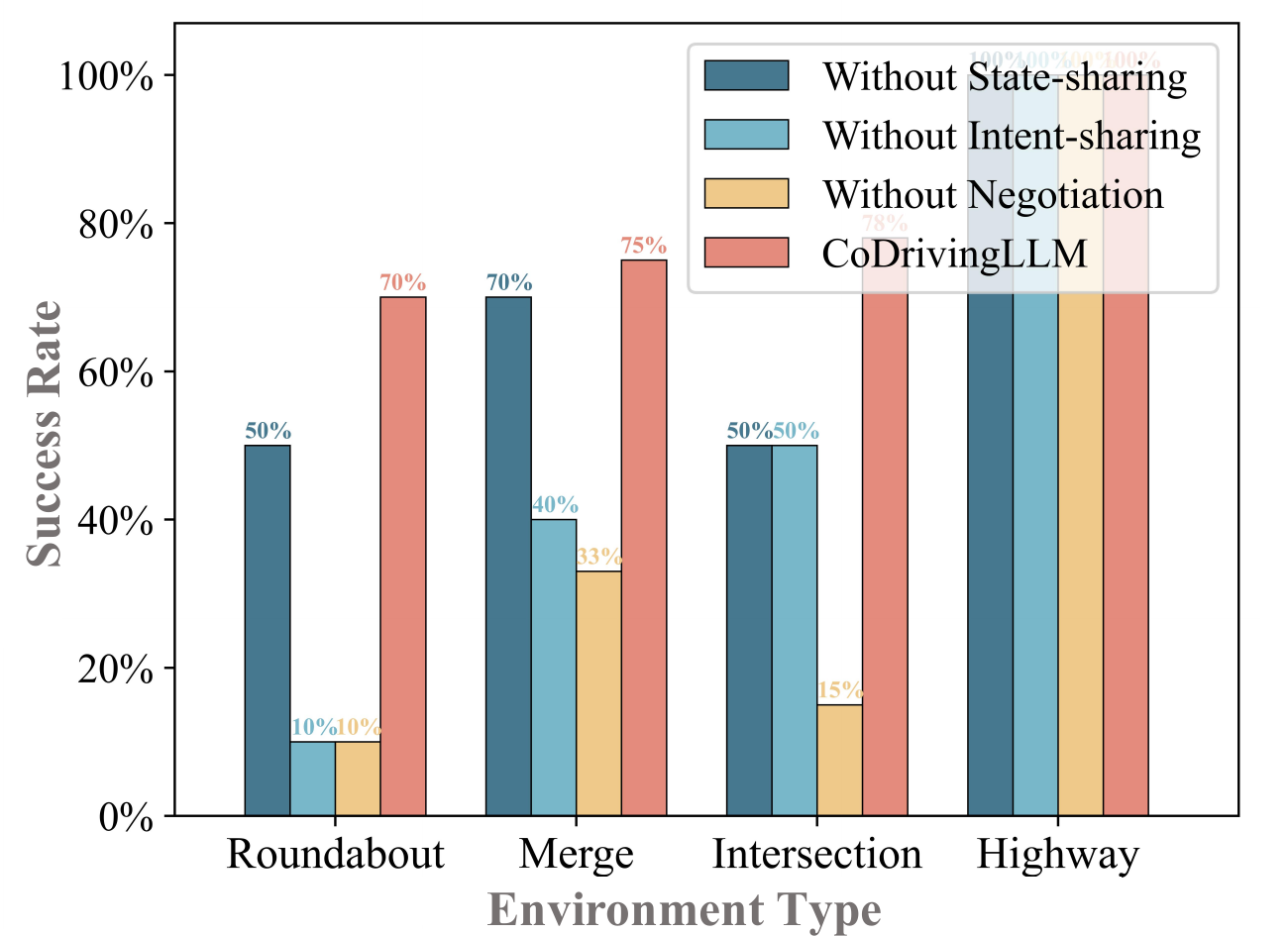}}
  \caption{The ablation experiment on sub-modules of the reasoning module.}\label{fig:ablation_negotiation}
  \end{center}
  \vspace{-0.8cm}
\end{figure}

As shown in Fig.\ref{fig:ablation_negotiation}, the negotiation module has the most significant impact on model performance. In roundabout and intersection, removing the negotiation module reduces the success rate to less than 15\%, which is less than one-fifth of the original model's performance. This result underscores the negotiation module's effectiveness in resolving complex traffic conflicts, making it a vital component of the cooperative driving framework. 

Additionally, the state-sharing module has a comparatively smaller effect. In all scenarios, the success rate remains above 50\%. It performs especially well in merging zones, where it closely matches the performance of CoDrivingLLM. This is due to the fact that vehicles in merging scenarios often travel parallel for extended stretches, allowing for earlier detection of interacting objects in adjacent lanes and more timely responses.

Finally, the intention-sharing module's impact varies across scenarios. In environments with concentrated conflict points, such as merging zones and intersections, its influence is relatively minor. However, in roundabouts, where conflict points are more dispersed and frequent, the absence of state sharing leads to a significant decline in success rates. Meanwhile, in highway scenarios, where car-following interactions dominate over conflicts, the success rate remains consistently high regardless of the module's presence.

\begin{figure*}[!htbp]
    \centering
    \includegraphics[width=7in]{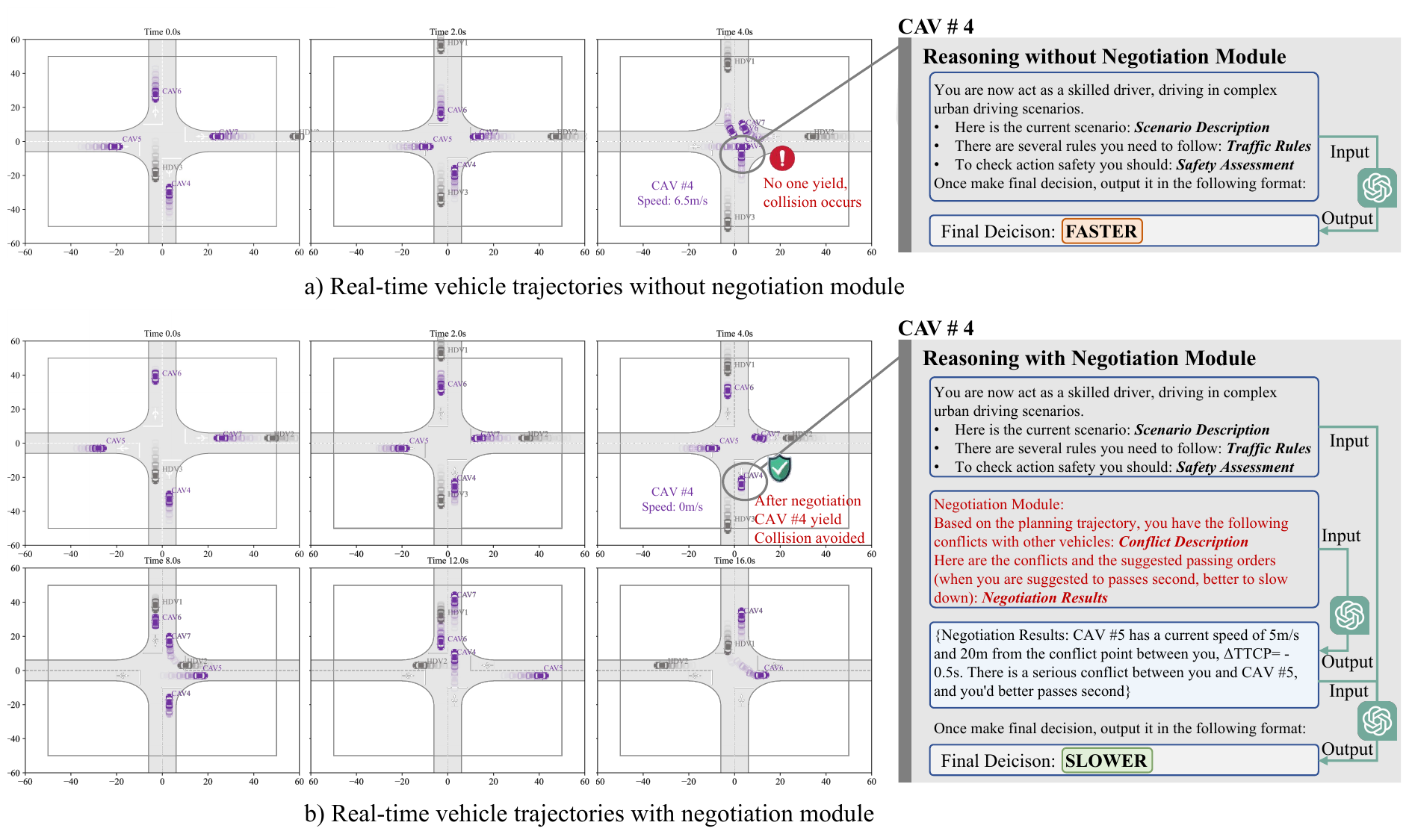}
    \caption{The cooperative driving cases at unsignalized intersection, (a) case from CoDrivingLLM without negotiation module, (b) case from CoDrivingLLM.}
    \label{fig:case_negotiation}
\end{figure*}

To better visualize the role of the negotiation module in the actual reasoning process, we selected a case with the same initial vehicle states in an intersection scenario to compare the different behaviors of vehicles with and without the negotiation module. Interaction cases videos and raw data of other scenarios can be found here\footnote{\url{https://fangshiyuu.github.io/CoDrivingLLM/}}.

In Fig.\ref{fig:case_negotiation}(a), CoDrivingLLM is not equipped with the negotiation module. At 2 seconds, the decision module fails to effectively predict potential interaction conflicts, and the lack of necessary coordination between CAVs leads to a situation at 4 seconds where the decision module cannot generate a safe and effective passing strategy. As a result, $\text{CAV}_4$ continues to accelerate, ultimately colliding with $\text{CAV}_5$. We can observe that in such complex conflict scenarios, even when LLMs are provided with clear scene information and safety assessment methods, they still cannot eliminate all interaction conflicts and traffic risks, which is unacceptable in real-world traffic.

In contrast, as shown in Fig.\ref{fig:case_negotiation}(b), CoDrivingLLM with the negotiation module effectively resolves this issue. In the same scenario, at 2 seconds, the negotiation module successfully identifies the potential movement conflict and, based on preset rules, suggests a recommended passing order, instructing $\text{CAV}_4$ to yield to $\text{CAV}_5$. This decision is ultimately adopted by the LLM in the decision module, leading $\text{CAV}_4$ to slow down at 4 seconds, allowing $\text{CAV}_5$ to pass safely, and successfully resolving the conflict.

This case demonstrates that our negotiation module effectively mitigates traffic conflicts and enhances the performance of the CoDrivingLLM method.

\subsection{The Performance of Continuous Learning}

The memory module enables CAVs to learn from past experiences, preventing the repetition of mistakes. To validate the continuous learning capability of the proposed framework, we compared the changes in the model's success rate after different numbers of interactions.
\begin{figure}[htbp]
  \begin{center}
  \centerline{\includegraphics[width=3.2in]{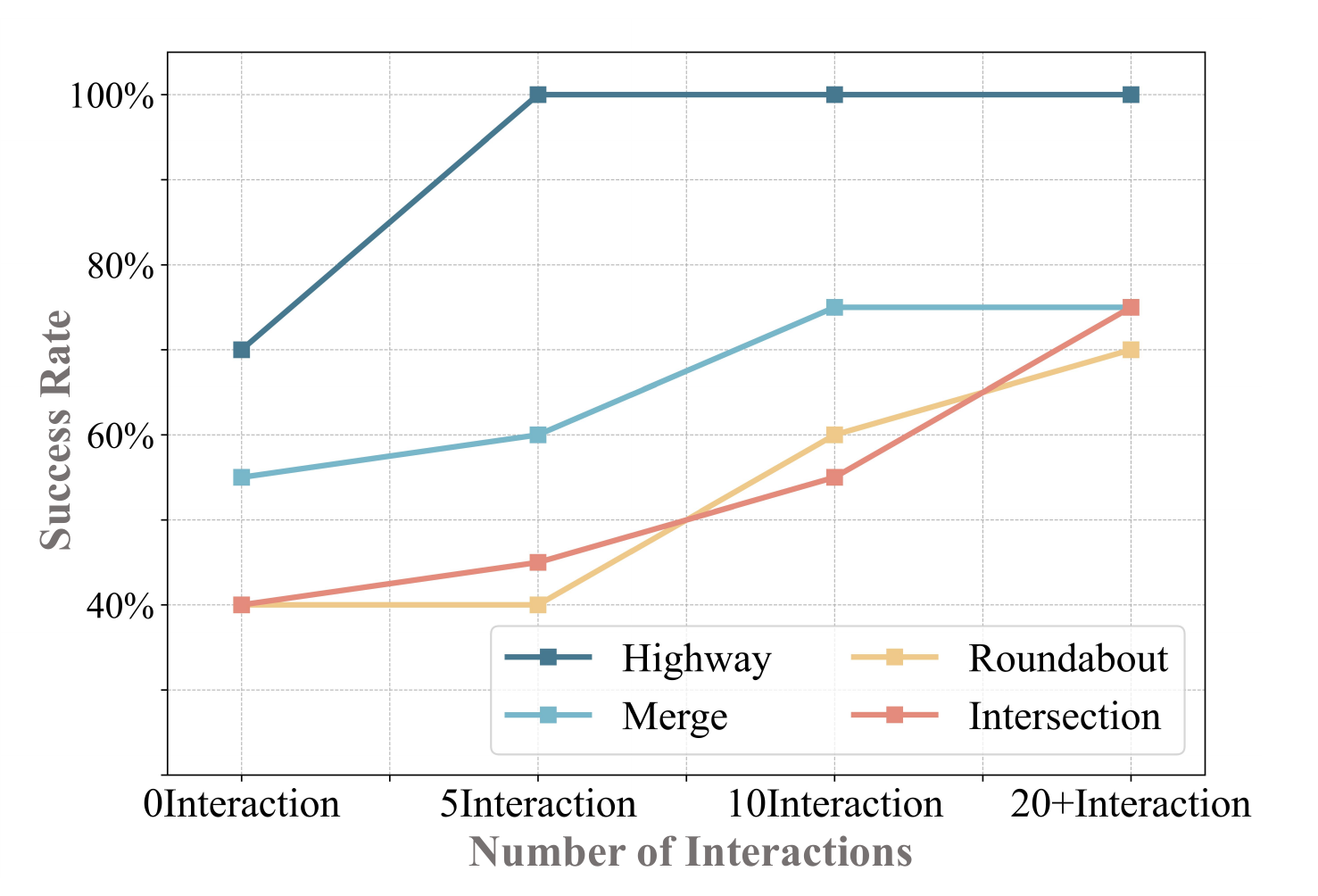}}
  \caption{The performance evaluation of memory module with different interaction numbers.}\label{fig:learning}
  \end{center}
  \vspace{-0.8cm}
\end{figure}

In the highway environment, as shown in Fig.\ref{fig:learning}, the success rate reaches 100\% after just 5 interactions. This high success rate can likely be attributed to the relatively simple conflict patterns and lower complexity compared to other scenarios. In the merging scenario, the success rate reaches around 75\% after 10 interactions. However, intersections and roundabouts show lower success rates during the initial interactions due to their higher complexity, which includes a variety of conflict types and multiple conflict points. Overall, the success rate improves significantly with the number of interactions across various scenarios, demonstrating that the memory module enables CoDrivingLLM to effectively learn from interactions.




\begin{figure*}[!htbp]
    \centering
    \includegraphics[width=7in]{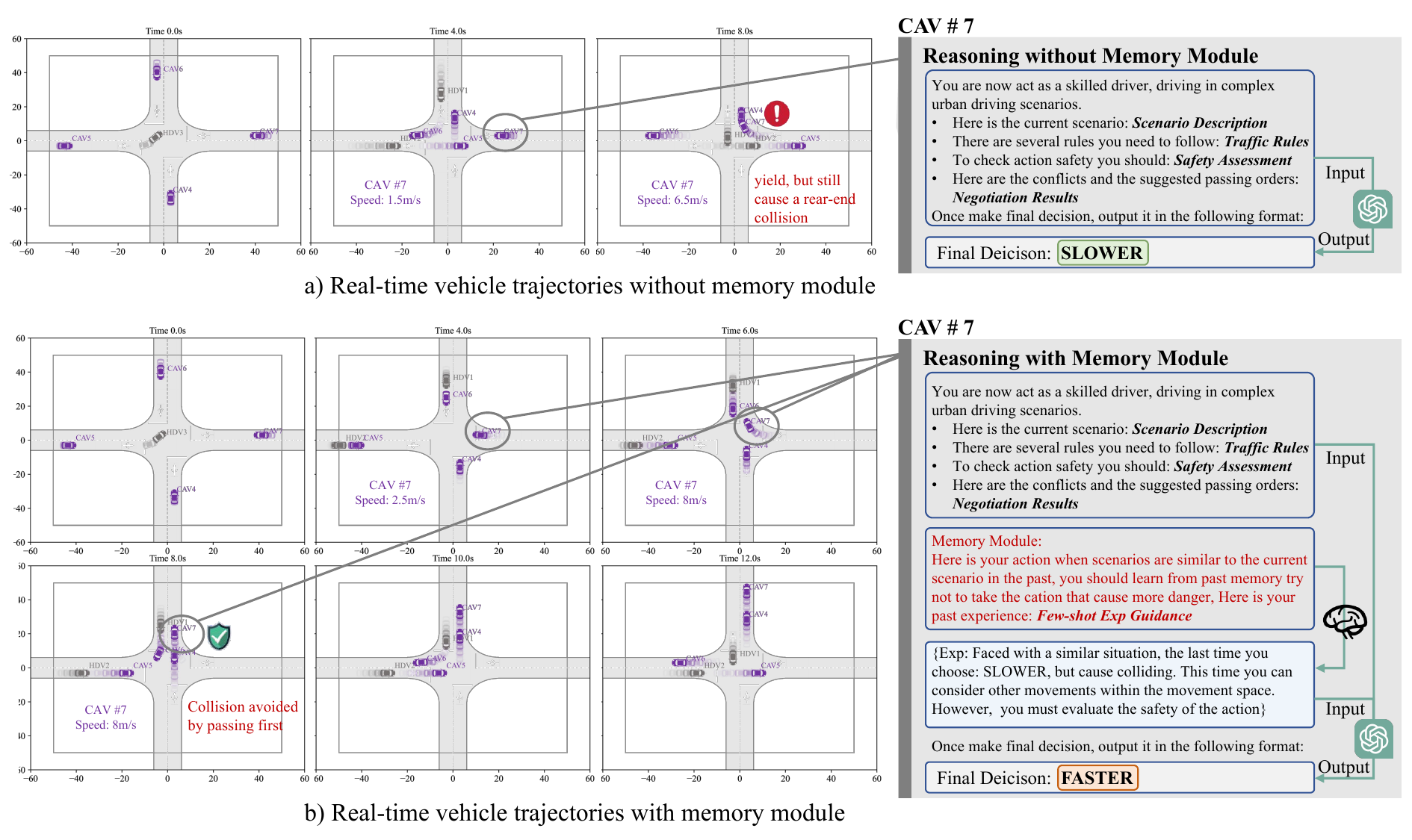}
    \caption{The cooperative driving cases at unsignalized intersection, (a) case from CoDrivingLLM without memory module, (b) case from CoDrivingLLM.}
    \label{fig:case_memory}
\end{figure*}

We also used an intersection scenario case to demonstrate the role of the memory module. Fig.\ref{fig:case_memory}(a) shows a typical case of decision failure. In this instance, CoDrivingLLM is not equipped with the memory module. At 4 seconds, although the decision module, through the negotiation module, identifies a potential rear-end conflict between $\text{CAV}_7$ and $\text{CAV}_4$ and attempts to resolve it by adjusting actions, $\text{CAV}_4$ accelerates while $\text{CAV}_7$ decelerates. However, due to $\text{CAV}_7$'s high initial speed, it collides with $\text{CAV}_4$ before it can fully decelerate. This case illustrates that in certain special scenarios, cooperative decision-making can still fail even with the negotiation module in place.

In contrast, Fig.\ref{fig:case_memory}(b) demonstrates the decision-making performance of CoDrivingLLM after utilizing the memory module. By retrieving memories, CoDrivingLLM identifies a similar scenario where a decision failed and conducts a few-shot experience learning session. At 4 seconds, it outputs a different decision strategy, instructing $\text{CAV}_7$ to accelerate and pass while $\text{CAV}_4$ slows down to yield. This decision is the opposite of the one in Fig.\ref{fig:case_memory}(a) and effectively resolves the conflict between the two CAVs, ultimately allowing all CAVs to reach their destinations safely.


\subsection{Comparison with Other Methods} 

Considering that not all cooperative driving methods are suitable for multiple scenarios, and unsignalized intersections are where accidents occur most frequently \cite{10397236}. Therefore, we selected the six typical cooperative driving methods mentioned earlier to test our approach under unsignalized intersection scenario, including optimization-based method (Cooperative game \cite{fang2024cooperative}, MCTS \cite{10706988}), rule-based method (iDFST \cite{chen2022conflict}, FCFS), learning-based method (MADQN \cite{egorov2016multi}), and LLM-based (Dilu \cite{wen2023dilu}).

\begin{table*}[t]
    \caption{Comparison under various methods}
    \centering
    \centering
    \begin{tabular}{c|c|c c c|c c c}
    \hline
     \multirow{2}{*}{Methods} & \multirow{2}{*}{Success rate} & \multicolumn{3}{c}{Post Encroachment Time} & \multicolumn{3}{c}{Travel Velocity}\\
     \cline{3-8}
       &  & Average & Max & Min & Average & Max & Min \\
     \hline
     FCFS &  \textbf{90\%} & 18.1 & 45.4 & 1.4 & 2.7 & 7.9 & 0.6 \\
     iDFST &  85\% & 15.1 & 47.6 & 1.2 & 4.1 & 7.9 & 0.2 \\
     MCTS &  65\% & 11.8 & 48.4 & 1.2 & 4.3 & 7.9 & 0.6 \\
     Cooperative game & 55\% & 5.7 & 18.6 & 0.8 & 5.7 & 7.9 & 0.9 \\
     MADQN  & 20\% & 3.7 & 9 & $\boldsymbol{1.8*}$ & $\boldsymbol{6.1*}$ & 6.5 & 5.4 \\
     Dilu & 38\% & 2.4 & 7.6 & 0.2 & 2.5 & 5.8 & 0.3 \\
     CoDrivingLLM (Ours) & \textbf{90\%} & 10.3 & 44 & $\boldsymbol{1.8*}$ & 5.7 & 8 & 1 \\
    \hline
\end{tabular}
    \label{tab:performance}
\end{table*}

\subsubsection{Overall Performance}
Firstly, we evaluated the overall performance of different methods based on their success rates, as summarized in Table. \ref{tab:performance}. The proposed CoDrivingLLM and the rule-based method FCFS both achieve a 90\% success rate across cases with various initial states, the highest among all methods. However, our method demonstrates a significant advantage in efficiency compared to FCFS, which will be discussed in detail below.  

In addition, the rule-based method iDFST achieves a success rate of 85\%. Among optimization-based approaches, MCTS and Cooperative game achieves success rates of 65\% and 55\%, respectively, placing them in the mid-range of performance. Moreover, although Dilu is a representative work that employs LLM for decision-making in CAV, it was originally designed for highway environments and not for cooperative driving. As a result, when applied to intersection scenarios, its success rate drop to 38\%. Lastly, MADQN displays the poorest generalization and performance, with a success rate of only 20\%.

\begin{figure}[htbp]
  \begin{center}
  \centerline{\includegraphics[width=3.5in]{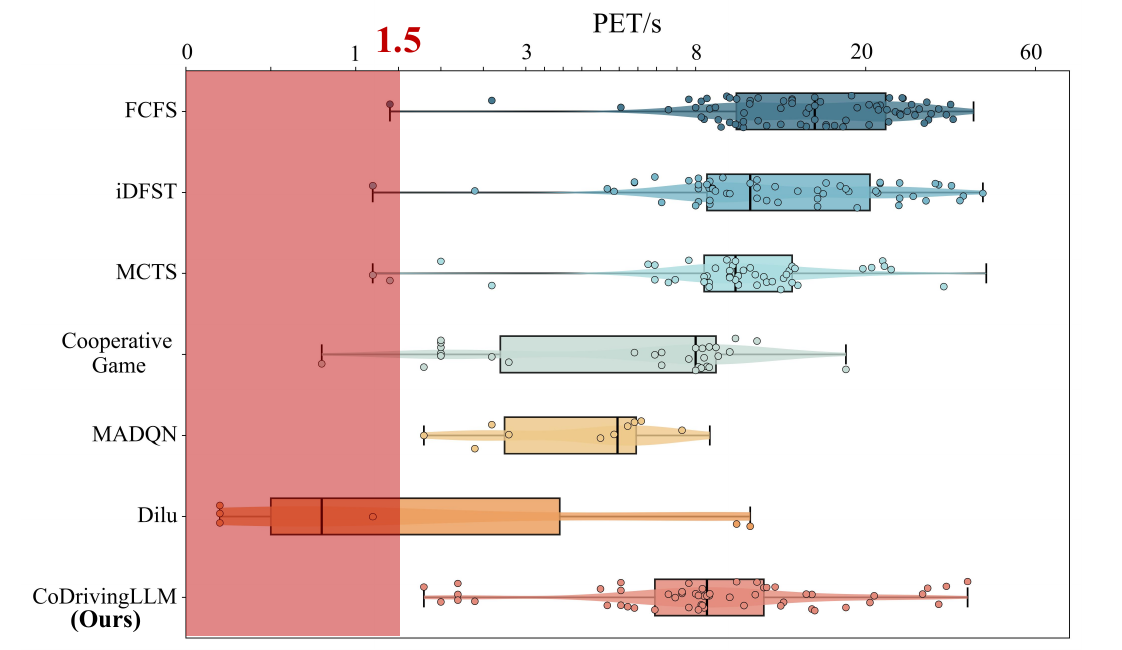}}
  \caption{PET evaluation results of different methods.}\label{fig:pet_result}
  \end{center}
  \vspace{-0.8cm}
\end{figure}

\subsubsection{Safety Evaluation}
Furthermore, we conducted a more in-depth analysis of the performance of different methods in terms of safety and efficiency. In this analysis, only successful cases were considered when calculating the relevant metrics. We have conducted a comprehensive safety analysis of various methods, utilizing the Post-Encroachment Time (PET) metric, a widely recognized safety parameter in traffic engineering. This metric effectively quantifies the safety and interaction intensity of vehicles in intricate traffic scenarios. According to the results presented in Fig. \ref{fig:pet_result} and Table. \ref{tab:performance}, the proposed CoDrivingLLM, FCFS, iDFST, and MCTS all show relatively high average PET values, indicating good stability in terms of safety. In contrast, Cooperative Game, MADQN, and Dilu exhibit average PET values of less than 6 seconds, suggesting that these methods involve a significant number of interactions with lower PET, which increases the risk of driving hazards.

Additionally, the minimum PET reflects the most dangerous interactions in collision-free cases, highlighting the critical moments when the risk of conflict is highest. Normally, a PET value less than 1.5s signifies severe conflicts that should be avoided \citep{2011Methodologies}. According to Fig. \ref{fig:pet_result}, The PET distribution of the proposed CoDrivingLLM is predominantly above 1.5s. Although the PET distribution for MADQN shows no values below 1.5s as well, considering that its success rate is only 20\%, there are a large number of collisions. Therefore, CoDrivingLLM performs best in terms of safety.

\subsubsection{Efficiency Evaluation}
We also evaluated the efficiency of all methods using travel velocity. As shown in Table. \ref{tab:performance}, MADQN achieves the highest average speed at 6.1 m/s during the tests. However, this efficiency comes at the cost of significantly compromising safety and overall performance. The proposed CoDrivingLLM achieves an average travel velocity of 5.7 m/s while ensuring safety, making it the most efficient method after excluding MADQN, which has a success rate of only 20\%. 

In addition, MCTS and Cooperative game demonstrate mid-range efficiency, partly due to their optimization objective, which includes minimizing delays. On the other hand, FCFS and iDFST, both rule-based methods, exhibit the lowest efficiency. Notably, while FCFS achieves the same overall success rate as CoDrivingLLM, its average travel velocity is only 2.7 m/s. Lastly, since Dilu is a single-vehicle decision-making method, it lacks  information sharing and negotiation. As a result, when applied to intersection scenarios, it leads to frequent deadlocks, resulting in poor efficiency.

We further compared the success rate of the proposed CoDrivingLLM with single-vehicle decision-making methods, available at here \footnote{\url{https://github.com/FanGShiYuu/CoDrivingLLM/blob/master/videos&data/Comparision.md}}. However, given that vehicle cooperation naturally provides an informational advantage through state and intent sharing, it is expected that CoDrivingLLM outperforms common single-vehicle decision-making methods. Therefore, we do not elaborate on this further.

In summary, CoDrivingLLM effectively balances safety and efficiency, achieving the best overall performance, which clearly demonstrates its superiority.




\section{Conclusion}
Cooperative driving technology effectively enhances the quality of cooperative decision-making for CAVs. Leveraging the rapid development of LLMs in recent years and their demonstrated strong logical reasoning and multi-task generalization capabilities, we propose an interactive, learnable LLM-driven cooperative decision-making framework. This framework includes three modules: the environment module, reasoning module, and memory module. Within this framework, we integrate four stages of cooperative decision-making, including state sharing, intent sharing, negotiation, and action based on different levels of CDA. Each CAV performs distributed high-level logical reasoning using LLMs, based on its own observations and the shared state and intent information from other vehicles, with centralized conflict resolution managed through a conflict coordinator within the framework. Additionally, we design a memory module to store decision-making processes and outcomes, enabling CAVs to achieve continuous learning and self-evolution through RAG technology.
Our method is tested across multiple cooperative scenarios, and the results demonstrate that our approach significantly outperforms other rule-based, optimization-based, and machine-learning methods in terms of success rate, safety, efficency.

In the future, we will continue to explore the potential applications of CoDriving in other scenarios, including its ability to handle heterogeneous background traffic. Additionally, we plan to conduct field tests to further validate its feasibility in more realistic interaction environments. Additionally, we intend to refine the memory module and evaluate various LLM base models to bolster the practical applicability and performance of our approach.

\ifCLASSOPTIONcaptionsoff
  \newpage
\fi

\footnotesize
\bibliographystyle{IEEEtranN}
\bibliography{IEEEabrv,Bibliography}

\vfill

\begin{IEEEbiography}[{\includegraphics[width=1in,height=1.25in,clip,keepaspectratio]{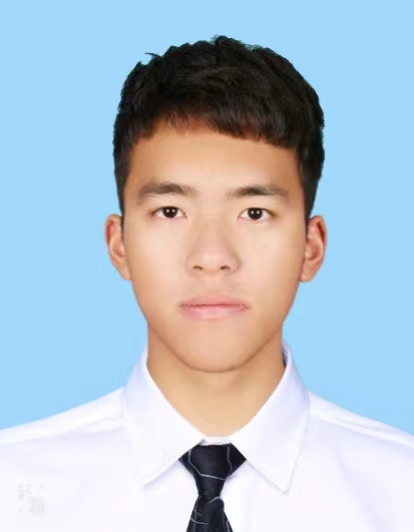}}]{Shiyu Fang}
received the B.S. degree in transportation engineering from Jilin University, Changchun,
China. He is currently pursuing the Ph.D. degree with the Department of Traffic Engineering, Tongji University, Shanghai, China. His main research interests include decision making and motion planning for autonomous vehicles.
\end{IEEEbiography}

\begin{IEEEbiography}[{\includegraphics[width=1in,height=1.25in,clip,keepaspectratio]{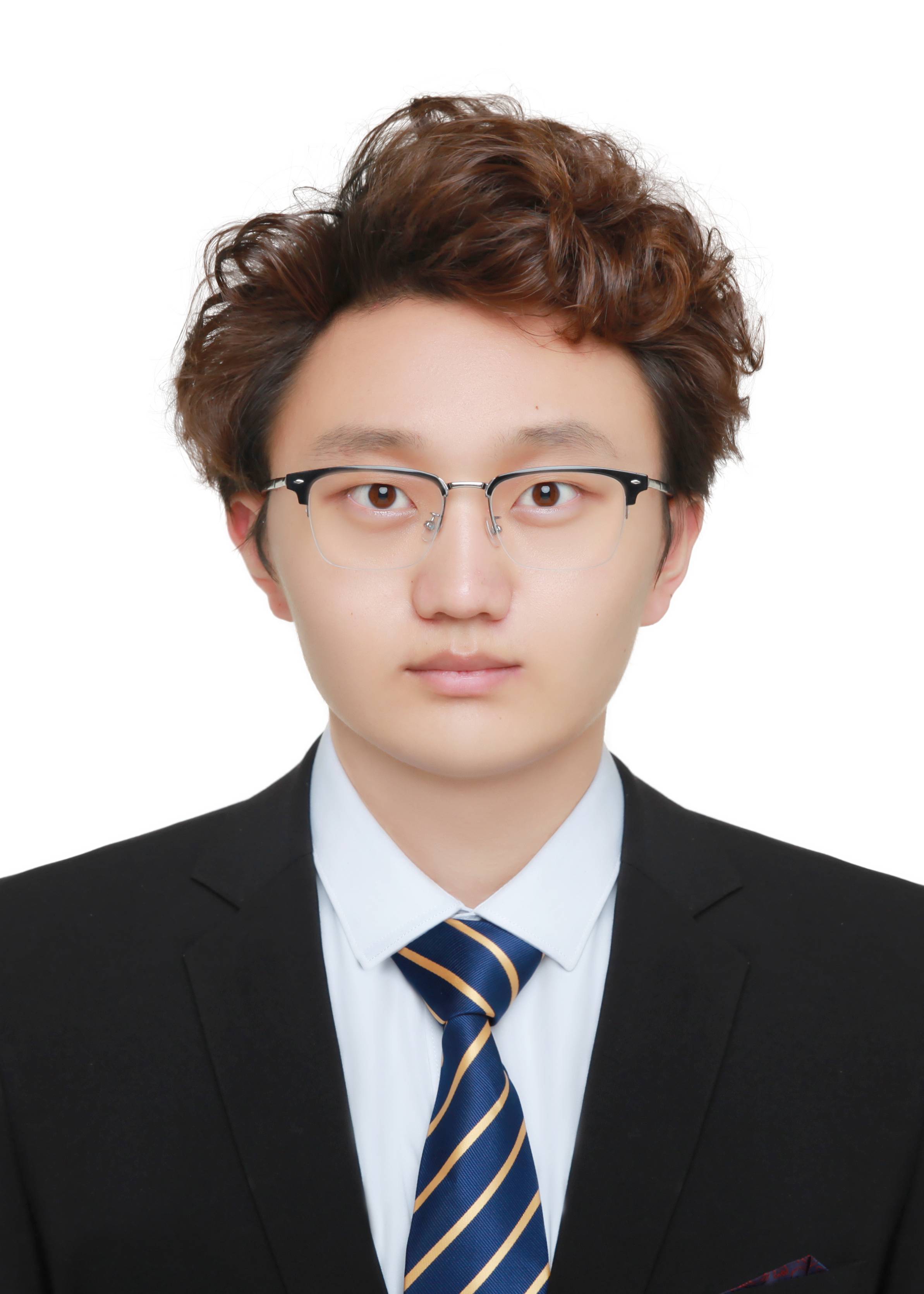}}]{Jiaqi Liu}
received the B.S. degree in transportation engineering from Tongji University, where he is currently pursing the M.S. degree. His research interests include decision-making of autonomous vehicles and data-driven traffic simulation. He won the Best Paper Award of CUMCM in 2020. He is a Visiting Researcher with the Department of Mechanical Engineering, University of California, Berkeley.
\end{IEEEbiography}

\begin{IEEEbiography}[{\includegraphics[width=1in,height=1.25in,clip,keepaspectratio]{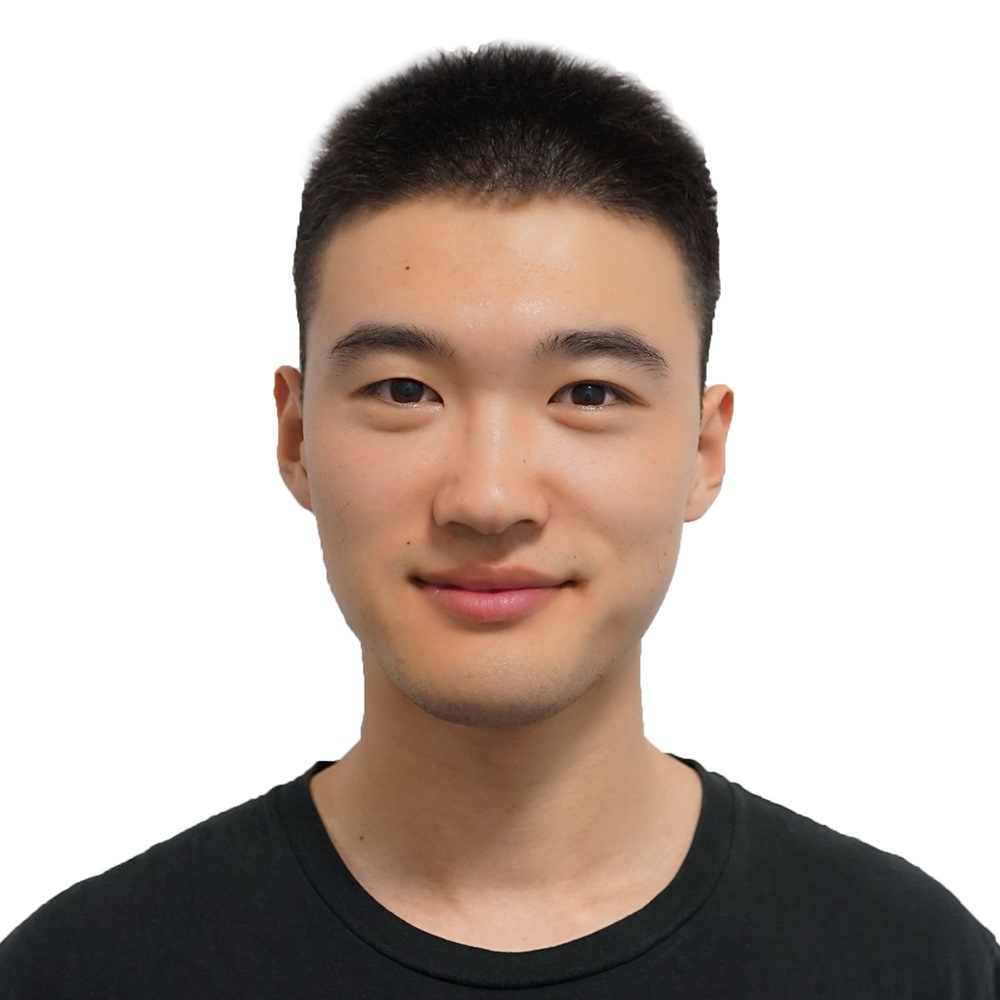}}]{Mingyu Ding}
 is an Assistant Professor in the Department of Computer Science at UNC-Chapel Hill and a postdoctoral fellow at UC Berkeley. Previously, he was a visiting scholar at MIT and earned his Ph.D. from the University of Hong Kong. Mingyu's research focuses on developing robots and embodied agents that can perceive, reason, and interact with the 3D physical world like humans, by integrating insights from robotics, computer vision, language-based reasoning, and brain science.
\end{IEEEbiography}

\begin{IEEEbiography}[{\includegraphics[width=1in,height=1.25in,clip,keepaspectratio]{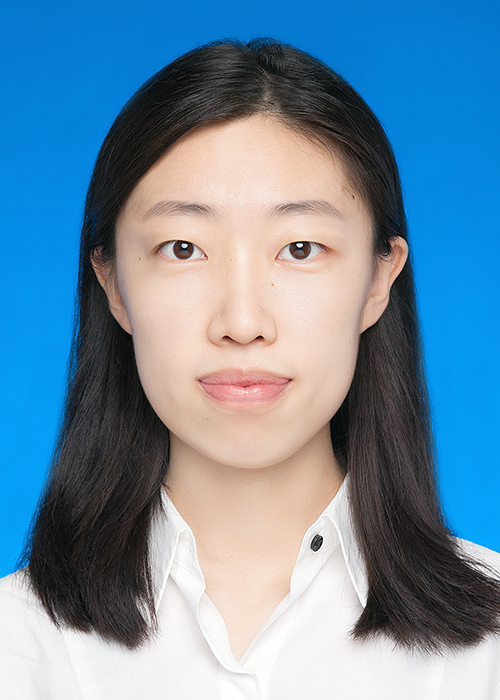}}]{Yiming Cui}
received the B.S. degree in transportation engineering from Tongji University, Shanghai,
China. She is currently pursuing the Ph.D. degree with the Department of Traffic Engineering, Tongji University, Shanghai, China. Her main research interests include decision making and motion planning for autonomous vehicles in the context of vehicle-infrastructure and vehicle-vehicle collaboration.
\end{IEEEbiography}

\begin{IEEEbiography}[{\includegraphics[width=1in,height=1.25in,clip,keepaspectratio]{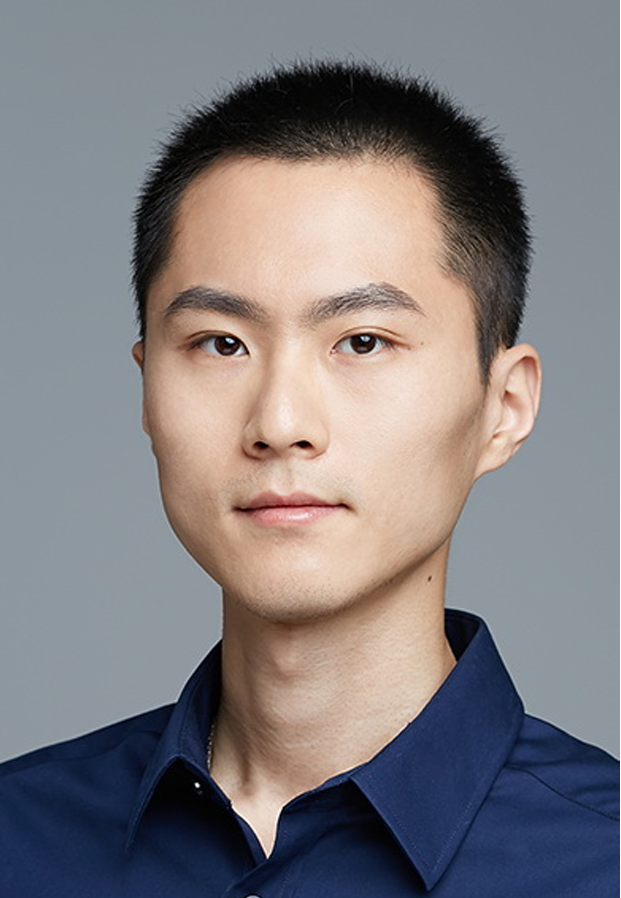}}]
{Chen Lv}  received his Ph.D. degree from the De
partment of Automotive Engineering, Tsinghua
 University, China, in 2016. From 2014 to 2015,
 he was a joint Ph.D. researcher in the EECS
 Dept., University of California, Berkeley.
 He is currently an Associate Professor at
 Nanyang Technology University, Singapore. His
 research focuses on advanced vehicles and
 human-machine systems, where he has con
tributed over 100 papers and obtained 12
 granted patents in China.
 Dr. Lv serves as an Associate Editor for IEEE TITS, IEEE T-VT, IEEE
 T-IV, and a Guest Editor for IEEE ITS Magazine, IEEE-ASME TMECH,
 Applied Energy, etc. He received many awards and honors, selectively
 including the Highly Commended Paper Award of IMechE UK in 2012,
 Japan NSK Outstanding Mechanical Engineering Paper Award in 2014,
 Tsinghua University Outstanding Doctoral Thesis Award in 2016, IEEE
 IV Best Workshop/Special Session Paper Award in 2018, Automotive
 Innovation Best Paper Award in 2020, the winner of Waymo Open
 Dataset Challenges at CVPR 2021, and Machines Young Investigator
 Award in 2022
\end{IEEEbiography}

\begin{IEEEbiography}[{\includegraphics[width=1in,height=1.25in,clip,keepaspectratio]{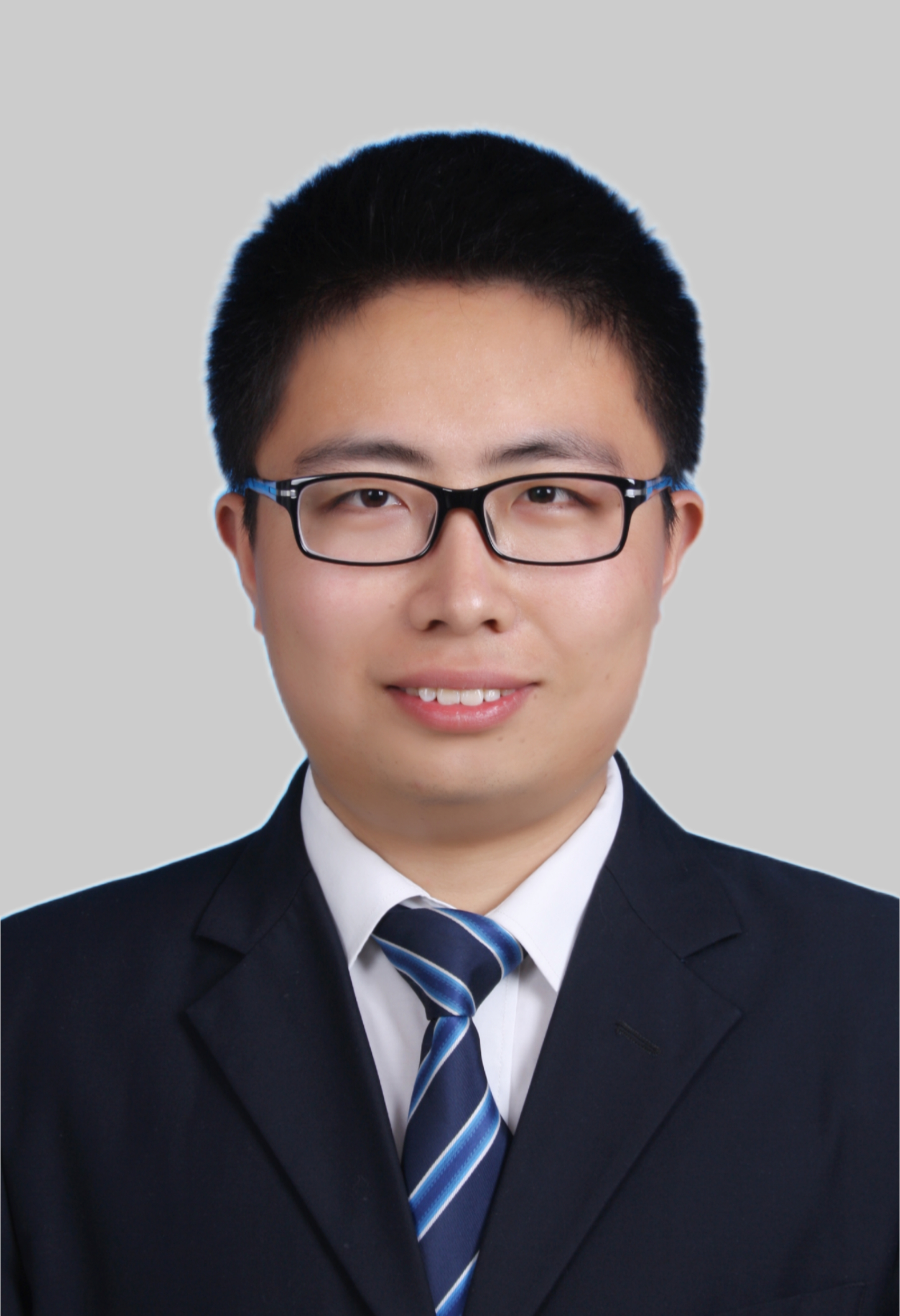}}]
{Peng Hang} is a Research Professor at the Department of Traffic Engineering, Tongji University, Shanghai, China. He received the Ph.D. degree with the School of Automotive Studies, Tongji University, Shanghai, China, in 2019. He was a Visiting Researcher with the Department of Electrical and Computer Engineering, National University of Singapore, Singapore, in 2018. From 2020 to 2022, he served as a Research Fellow with the School of Mechanical and Aerospace Engineering, Nanyang Technological University, Singapore. His research interests include vehicle dynamics and control, decision making, motion planning and motion control for autonomous vehicles. He serves as an Associate Editor of IEEE Transactions on Vehicular Technology, Journal of Field Robotics, IET Smart Cities, and SAE International Journal of Vehicle Dynamics, Stability, and NVH.
\end{IEEEbiography}

\begin{IEEEbiography}[{\includegraphics[width=1in,height=1.25in,clip,keepaspectratio]{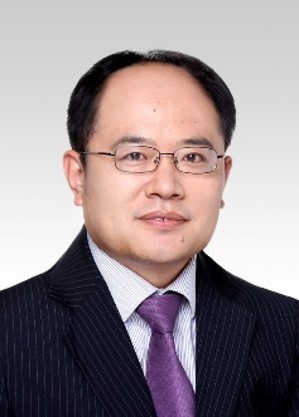}}]
{Jian Sun} received the Ph.D. degree from Tongji University in 2006. Subsequently, he was at Tongji University as a Lecturer, and then promoted to the position as a Professor in 2011, where he is currently a Professor with the College of Transportation Engineering and the Dean of the Department of Traffic Engineering. His main research interests include traffic flow theory, traffic simulation, connected vehicle-infrastructure system, and intelligent transportation systems.
\end{IEEEbiography}

\end{document}